\documentclass[runningheads]{llncs}

\usepackage[T1]{fontenc}
\usepackage{graphicx}
\usepackage{amsmath}
\usepackage{amssymb}
\usepackage{booktabs}
\usepackage{url}
\begin{document}

\title{AdaptReg: Skip-level FiLM Adapter for Task-Specific Adaptation of a Frozen Registration Foundation Model}
\titlerunning{AdaptReg}

\author{Shuyue Shi\inst{1}\thanks{Corresponding author.} \and Rong Yuan\inst{1} \and
Lin Tian\inst{2}}
\authorrunning{S. Shi et al.}
\institute{Shanghai Medical Image Insights Intelligent Technology Co., Ltd. \and
Hong Kong Baptist University}

\maketitle

\begin{abstract}
Longitudinal PSMA-PET/CT registration is a key step in assessing treatment
response in prostate cancer, and it requires more than generic alignment
accuracy: the deformation must preserve lesion volume, so that measured changes
in metabolic tumor volume (MTV) and total lesion glycolysis (TLG) reflect
treatment response rather than registration artifacts. We propose AdaptReg,
which adapts a frozen registration foundation model to such task-specific
requirements through a lightweight trainable adapter branch with
identity-preserving
initialization. Keeping uniGradICON fully frozen, we apply FiLM-style per-voxel
affine modulation ($\gamma$, $\beta$) to its decoder skip connections; the
modulation is generated from anatomical semantic features of a frozen
TotalSegmentator encoder and supervised by a lesion volume-preservation loss.
On the official Codabench
validation set (20 pairs), AdaptReg achieves DSC 0.7585, HD95 6.33\,mm,
MTV/TLG errors of 1.68\%/2.11\%, and 0.0022\% non-diffeomorphic voxels,
improving on the
uniGradICON + instance optimization baseline by $+0.05$ DSC and reducing the
MTV/TLG errors by 70.9\%/67.7\%. On the hidden official test set (160 pairs),
AdaptReg achieves CT DSC 0.816, PET DSC 0.598, and MTV/TLG errors of
4.02\%/3.46\%, ranking fourth among 16 teams in the final PSMAReg evaluation.
Code: \url{https://github.com/shishuyue/AdaptReg}.

\end{abstract}

\section{Introduction}

PSMA-PET/CT is central to staging and treatment response assessment in prostate
cancer, and deformably registering the follow-up scan to the baseline enables
quantification of changes in metabolic tumor volume (MTV) and total lesion
glycolysis (TLG). The Learn2Reg 2026 PSMAReg
challenge~\cite{learn2reg2026psmareg}, the latest edition of the Learn2Reg
registration challenge series~\cite{hering2022learn2reg,hansen2025learn2reg},
targets this task: with the baseline fixed
and the follow-up moving, predict the dense displacement field, evaluated by the
Dice of 117 warped organ labels (DSC), HD95, lesion MTV/TLG percentage errors,
and the non-diffeomorphic voxel ratio (\%NDV).

Inverse-consistency regularization, introduced by ICON~\cite{greer2021icon} and
refined by GradICON~\cite{tian2023gradicon}, underpins
uniGradICON~\cite{tian2024unigradicon}, a registration foundation model with
strong zero-shot generalization. Such a model, however, is trained for generic
alignment quality and offers no mechanism for enforcing task-specific
constraints such as the lesion volume preservation that the MTV/TLG metrics
demand.

\begin{figure}[t]
\centering
\includegraphics[width=\linewidth]{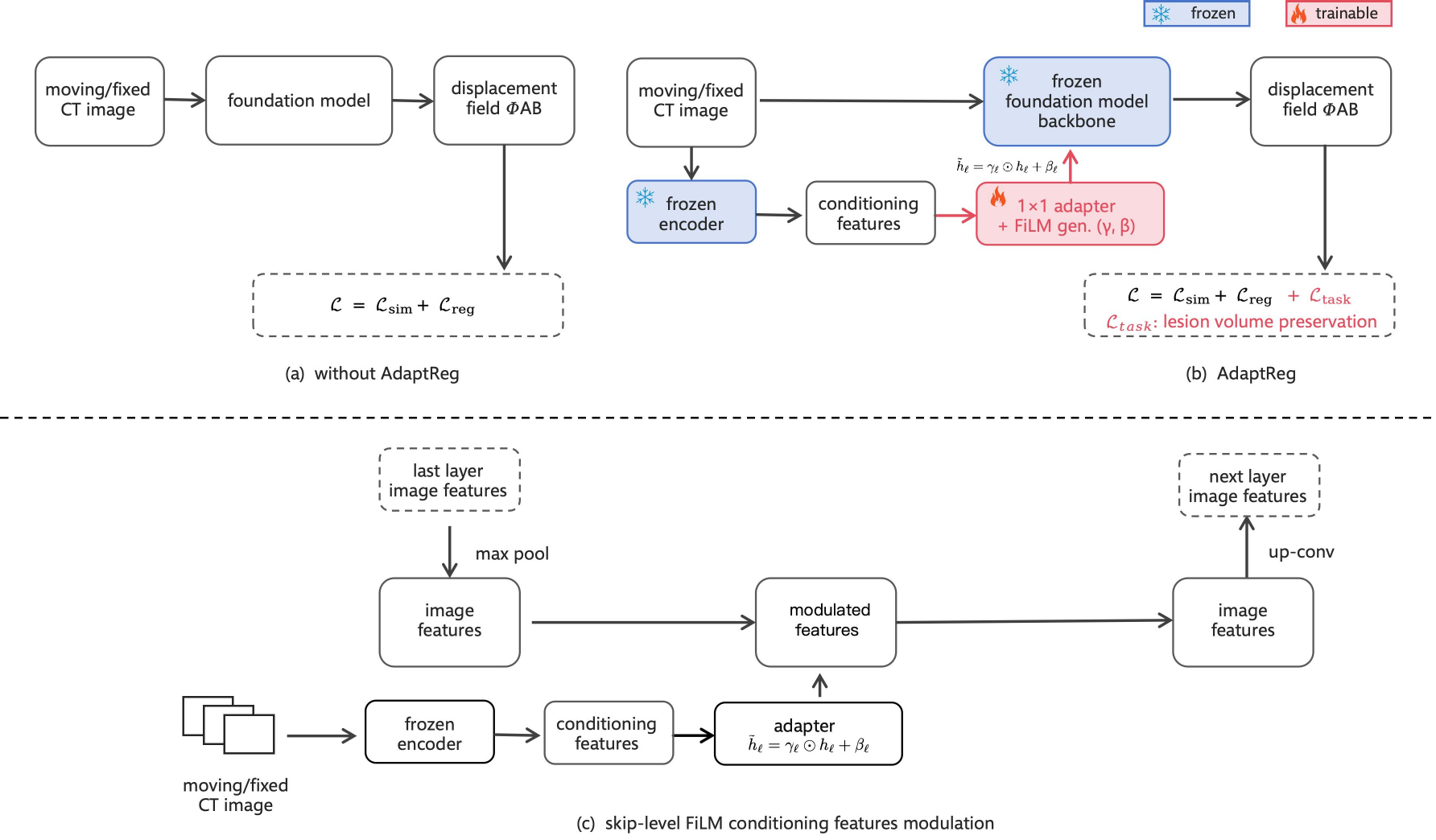}
\caption{(a) Without AdaptReg, the foundation model is trained with similarity
and regularization losses only. (b) AdaptReg adds an adapter branch
(frozen encoder + trainable FiLM generator modulating the decoder
skips) and a task loss (lesion volume preservation). (c) Skip-level FiLM
modulation detail.}
\label{fig:arch}
\end{figure}

We therefore propose \textbf{AdaptReg} (Fig.~\ref{fig:arch}), which keeps the
foundation model frozen and steers it with a lightweight trainable branch,
initialized near identity so that training starts from exactly the backbone's
behavior. Specifically, we inject FiLM~\cite{perez2018film}-style
per-voxel
affine modulation into the decoder skip connections, which carry the
high-frequency features that determine the spatial detail of the displacement
field; we generate the modulation from anatomical semantic features of a frozen
segmentation network; and we supervise it with a lesion volume-preservation loss
(Sec.~\ref{sec:losses}) that serves the official MTV/TLG metrics.

\section{Method}

\subsection{Frozen Backbone and Adapter Branch}
\label{sec:backbone}

Given a moving image $I^A$ and a fixed image $I^B$, a registration model
predicts the transformation $\Phi^{AB}$ that maps $I^A$ onto $I^B$. We register
the CT volumes; PET and CT are acquired in alignment, so the predicted field
transfers to the PET and lesion masks annotated on PET are valid in CT space.
Instead of
training such a model from scratch, we control the transformation predicted by a
foundation model so that it additionally fulfills a task-specific requirement,
here lesion-region volume preservation. We use
uniGradICON~\cite{tian2024unigradicon}, a cascade of four U-Nets, as a fully frozen backbone and attach an
adapter branch~\cite{rebuffi2017adapters,houlsby2019adapter}
(Fig.~\ref{fig:arch}). Its conditioning features are the 32-channel activations
of the \emph{first} encoder stage of a frozen
TotalSegmentator~\cite{wasserthal2023totalsegmentator}
nnU-Net~\cite{isensee2021nnunet}, taken for both CTs at the resolution the
corresponding cascade stage operates on, concatenated, and
compressed to 16 channels by a trainable $1{\times}1$ convolution. All deeper
TotalSegmentator stages and its decoder are discarded, so no layer-wise
correspondence between the two architectures is required. The two paths
use different intensity preprocessing: the branch input follows
TotalSegmentator (HU clipped to $[-1024, 276]$ and z-scored), whereas the
backbone input retains uniGradICON's original preprocessing (HU clipped to
$[-1000, 1000]$ and min-max normalized to $[0, 1]$).

\subsection{Skip-level FiLM Modulation}
\label{sec:film}

At each of the four cascaded stages, the U-Net is replaced by a
modulation-capable copy initialized layer by layer from the official checkpoint,
with a dense FiLM layer before each decoder skip concatenation:
\begin{equation}
\tilde{h}_{\ell} = \gamma_{\ell} \odot h_{\ell} + \beta_{\ell} ,
\end{equation}
where $h_{\ell}$ is the encoder feature map carried by the $\ell$-th skip
connection and $\gamma_{\ell}$, $\beta_{\ell}$ are dense spatial maps of the same
shape as $h_{\ell}$. They are generated per stage from the conditioning features
by a branch-local pyramid: a $1{\times}1$ adapter fuses those features with
$h_{1}$ (in uniGradICON's U-Net, the concatenated input pair), five convolutional
blocks with max pooling then walk down to each decoder resolution, and a
per-level $1{\times}1$ convolution projects to that level's $\gamma_{\ell}$ and
$\beta_{\ell}$. The pyramid's depth and per-level output channel counts are the
only quantities read off the backbone ($5$ levels, $2/16/32/64/256$ channels);
where rounding makes a level's size differ from $h_{\ell}$, the parameters are
trilinearly resampled. Both are near-identity
initialized ($\gamma$ bias 1, $\beta$ bias 0, $\gamma/\beta$ convolution weights
rescaled by $10^{-2}$). We keep these weights small-random rather than exactly
zero: a zero-initialized head would block gradient flow into the conditioning
encoder, whereas a $10^{-2}$ scale perturbs the frozen backbone's forward pass by
only $\sim$1\% while letting the branch receive gradient from the first step.

\subsection{Training Losses}
\label{sec:losses}

The training loss combines a multi-class Dice term $L_{\mathrm{dice}}$, the
GradICON regularizer $L_{\mathrm{GIC}}$, and a lesion volume-preservation term
$L_{\mathrm{vol}}$,
\begin{equation}
L = 4 \cdot L_{\mathrm{dice}} + 0.1 \cdot \lambda(t;T_{\mathrm{GIC}}) \cdot L_{\mathrm{GIC}} + 6 \cdot \lambda(t;T_{\mathrm{vol}}) \cdot L_{\mathrm{vol}},
\end{equation}
with a linear warmup $\lambda(t;T) = \min(1,\, t/T)$ ramping in the two
regularizers over the first $T$ epochs ($t$: current epoch,
$T_{\mathrm{GIC}} = 25$, $T_{\mathrm{vol}} = 50$). $L_{\mathrm{dice}}$, the only
matching-driven term, is the soft Dice loss between the fixed labels and the
moving labels warped by $\Phi^{AB}$, averaged over the classes present in the
fixed image and computed with linear interpolation so that gradients reach the
displacement field. For regularization we adopt the GradICON~\cite{tian2023gradicon} term
used to train uniGradICON~\cite{tian2024unigradicon}, $L_{\mathrm{GIC}} = \|
\nabla(\Phi^{AB} \circ \Phi^{BA}) - \mathrm{Id} \|_F^2$, with $\mathrm{Id}$ the
identity and $\Phi^{BA}$ the reverse-direction prediction of the same network;
the term suppresses folding and promotes diffeomorphic maps. $\Phi^{BA}$ is a
second full forward pass with the two inputs swapped, including recomputation
of the conditioning features in the swapped order; it is evaluated without
gradient tracking, so $L_{\mathrm{GIC}}$ backpropagates only through
$\Phi^{AB}$.

To satisfy the task-specific requirement on the transformation, we introduce a
lesion volume-preservation loss
\begin{equation}
L_{\mathrm{vol}} = \sum_x m^B(x)\,(\det J(x) - 1)^2 \Big/\sum_x m^B(x),
\end{equation}
where $J(x) = \nabla \Phi^{AB}(x)$ is the Jacobian of the transformation in the
fixed image space and $m^B = \mathbf{1}[M^A_{\mathrm{lesion}} > 0] \circ
\Phi^{AB}$ is the moving lesion mask warped to that space. $\det J$ is computed
per voxel from the predicted displacement field by centered finite differences
(one-sided at the volume boundary) followed by the analytic $3{\times}3$
determinant. Driving $\det J$
toward 1 inside
the lesion keeps its voxel count (MTV) and SUV integral (TLG) unchanged under
warping. Since $m^B$ depends on $\Phi^{AB}$, the loss admits a degenerate
solution in which the network shrinks the warped mask until the weighted average
becomes vacuous; we therefore detach $m^B$, so that gradients reach the
transformation only through $\det J$. For training pairs without a lesion
annotation, $m^B$ is empty and the term is skipped (defined as zero), so the
normalization is never vacuous.

\section{Experiments and Results}

\subsection{Experimental Setup}

\textbf{Data and preprocessing.} The challenge training images are derived
from the public PSMA-PET-CT-Lesions dataset~\cite{jeblick2024psma} (597 PSMA
PET/CT studies acquired at LMU University Hospital Munich, with manually
annotated tumor lesions). We train at the native volume shape,
$288{\times}192{\times}192$, without resampling; for comparison we also train a
model at the $175^3$ shape originally used by uniGradICON. The paired training
set comprises 146 patients, each with one baseline scan and at least one
follow-up: 114 patients have a single follow-up and 32 have two or more (up to
six), which yields 202 baseline$\to$follow-up pairs in total. For every patient
with $n \ge 2$ follow-ups we additionally form all $\binom{n}{2}$
follow-up$\to$follow-up pairs, contributing 97 further pairs, for 299 pairs in
total. The split is made at the patient level so that no patient appears in
both subsets: 134 patients (278 pairs) for training and 12 patients (21 pairs)
for validation. The anatomical labels (up to 117 classes per volume) are
provided by the organizers with the training cases; $L_{\mathrm{dice}}$ is
computed only over the classes present in each pair (typically 115--116), with
the per-class one-hot masks processed in chunks of 32 classes to bound GPU
memory. Augmentation comprises
random fixed/moving swapping, synchronized 3D rotation ($\pm 10^{\circ}$) and
translation ($\pm 10$ voxels), and HU jittering ($\pm 50$), each with
probability 0.5.

\textbf{Training.} We train in a single 500-epoch run with Adam (weight decay
$10^{-4}$, learning rate $10^{-3}$ cosine-decayed to $10^{-5}$) and DDP on four
RTX~4090 GPUs, with batch size 1 per GPU (effective batch 4, $\sim$280 steps
per epoch), for a total training time of about 34\,h; activation checkpointing keeps
native-shape training within 24\,GB per GPU. Checkpoints are selected by the
validation CT-Dice loss, which correlates almost perfectly with the official DSC
($r = -0.994$).

\textbf{Inference.} An ANTs rigid pre-registration is estimated on the CTs
(Mattes mutual information with 64 histogram bins, computed on bed-removed,
$[-300, 300]$\,HU-windowed, $2{\times}$ downsampled images, seeded and
single-threaded for determinism), and
the moving CT is warped by the resulting field $u_{\mathrm{rig}}$ before
entering the network; the network's residual field is composed with it as
$u(q) = u_{\mathrm{def}}(q) + u_{\mathrm{rig}}(q + u_{\mathrm{def}}(q))$.
Residual edge artifacts from padding are removed by zeroing the field at voxels
whose displacement magnitude exceeds half the image size along any axis. The
PET volumes are accepted per the official interface but do not affect the
predicted field. The
pipeline is deterministic and ships as the submitted Docker image. Code and the
submitted container are available at \url{https://github.com/shishuyue/AdaptReg}.

\subsection{Results}

\begin{table}[t]
\centering
\caption{Results on the official validation set (20 pairs), mean $\pm$ std.
HD95 in mm; MTV, TLG, and \%NDV in percent.}
\label{tab:main}
\scriptsize
\setlength{\tabcolsep}{2.5pt}
\renewcommand{\arraystretch}{2}
\begin{tabular}{@{}lccccc@{}}
\toprule
Method & DSC $\uparrow$ & HD95 $\downarrow$ & MTV $\downarrow$ & TLG $\downarrow$ & \%NDV $\downarrow$ \\
\midrule
\shortstack[l]{uniGradICON\\+ IO (50 iters.)} & $0.7089\,{\pm}\,0.0377$ & $6.784\,{\pm}\,2.156$ & $5.77\,{\pm}\,3.81$ & $6.54\,{\pm}\,4.18$ & $0.0017\,{\pm}\,0.0051$ \\ \addlinespace[8pt]
\shortstack[l]{AdaptReg\\($175^3$ training)} & $0.7245\,{\pm}\,0.0403$ & $6.728\,{\pm}\,2.504$ & $1.91\,{\pm}\,2.47$ & $\mathbf{2.09\,{\pm}\,2.59}$ & $\mathbf{1.5{\times}10^{-5}\,{\pm}\,5{\times}10^{-5}}$ \\ \addlinespace[8pt]
\shortstack[l]{AdaptReg\\(native shape)} & $\mathbf{0.7585\,{\pm}\,0.0370}$ & $\mathbf{6.331\,{\pm}\,2.321}$ & $\mathbf{1.68\,{\pm}\,1.87}$ & $2.11\,{\pm}\,2.34$ & $0.0022\,{\pm}\,0.0018$ \\
\bottomrule
\end{tabular}
\end{table}

\begin{figure}[t]
\centering
\includegraphics[width=\linewidth]{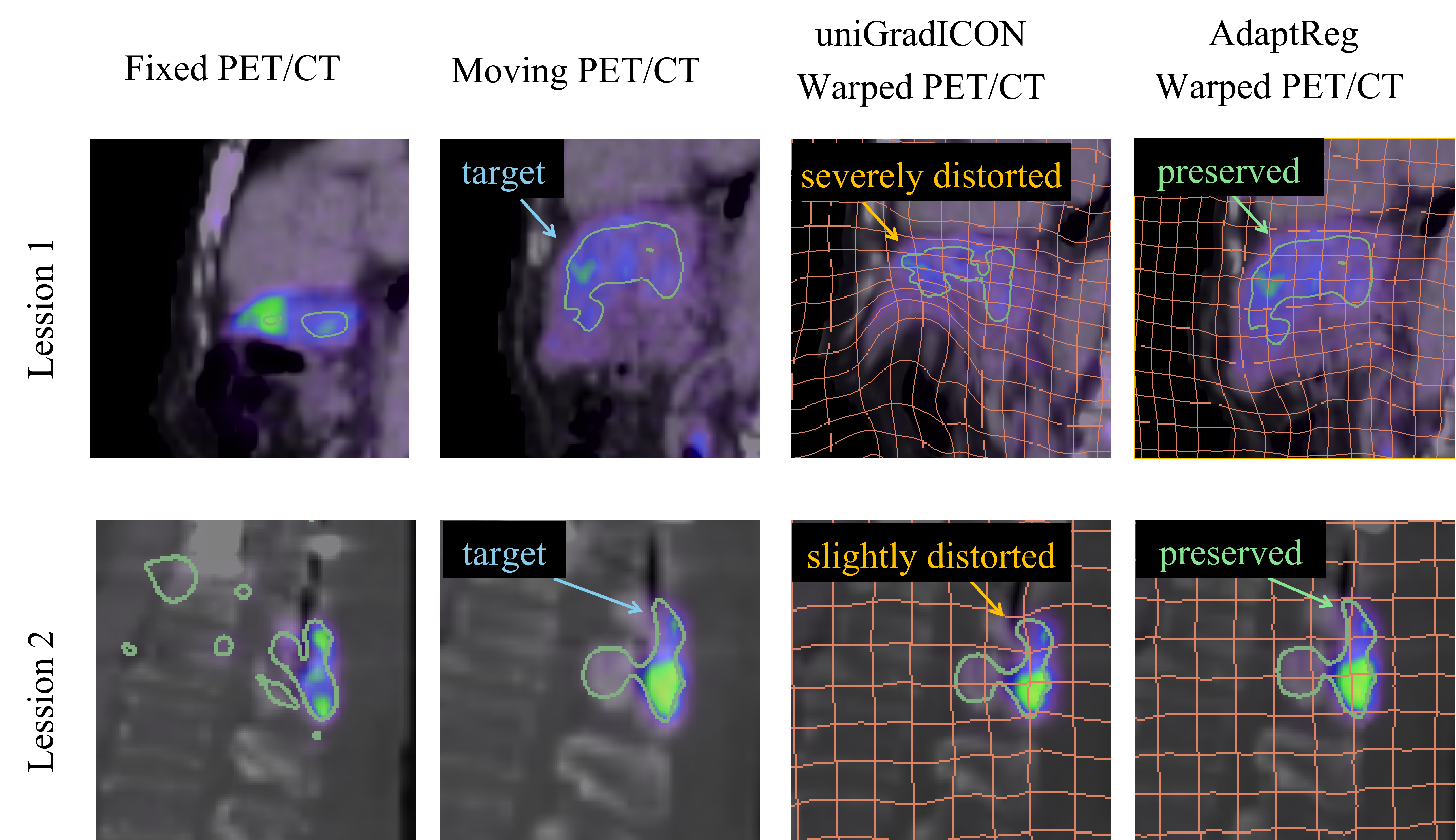}
\caption{Qualitative comparison on two validation lesions (sagittal PET/CT
overlay with the deformation grid). Columns: fixed, moving, uniGradICON-warped,
and AdaptReg-warped PET/CT. The generic backbone visibly distorts the lesion
(outlined), whereas AdaptReg preserves its shape, consistent with the lower
MTV/TLG errors in Table~\ref{tab:main}.}
\label{fig:qual}
\end{figure}

Table~\ref{tab:main} compares AdaptReg with the uniGradICON + instance
optimization (IO) baseline on the official validation set (labels hidden); the
IO baseline fine-tunes all uniGradICON parameters per pair for 50 iterations
with Adam (learning rate $2{\times}10^{-5}$) under the default uniGradICON
objective (LNCC similarity with $\sigma{=}5$ plus the GradICON
inverse-consistency term of weight 1.5).
The $175^3$-trained model already surpasses the baseline on MTV/TLG, though only
marginally on DSC; training at the native shape realizes the full gain of
$0.0496$ DSC ($0.7089 \to 0.7585$) and relative MTV/TLG error reductions of
70.9\%/67.7\%. Comparing
the two, resampling to $175^3$ distorts small anatomical structures or shrinks
them to a negligible size, lowering their Dice and accounting for much of the
gap. The rest follows the design: MTV and TLG respond to $L_{\mathrm{vol}}$
(Fig.~\ref{fig:qual}), whereas the DSC and HD95 gains come from the anatomical
semantic conditioning and the native shape. On the validation
set, regularity is retained, with \%NDV well below the official
penalty threshold of 0.005\%.

\begin{table}[t]
\centering
\caption{Ablation on the official validation set (20 pairs), mean over pairs.
Top: without vs.\ with $L_{\mathrm{vol}}$ at the $175^3$ shape (150-epoch
runs). Bottom: the weight of $L_{\mathrm{vol}}$ and the inference pipeline at
the native shape (500-epoch runs). naive: plain half-resolution submission;
the $w{=}4$ row is its best-Dice checkpoint, and the $w{=}6$ rows share the
submitted checkpoint. LS: least-squares pre-compensation of the
scorer's fixed $2{\times}$ upsampling, a step specific to the validation-phase
half-resolution submission format and not used by the full-resolution
test-phase Docker.}
\label{tab:ablation}
\scriptsize
\setlength{\tabcolsep}{2.5pt}
\renewcommand{\arraystretch}{1.6}
\begin{tabular}{@{}lccccc@{}}
\toprule
Configuration & DSC $\uparrow$ & HD95 $\downarrow$ & MTV $\downarrow$ & TLG $\downarrow$ & \%NDV $\downarrow$ \\
\midrule
$175^3$, w/o $L_{\mathrm{vol}}$ & 0.7259 & 6.801 & 6.61 & 9.37 & $1.6{\times}10^{-3}$ \\
$175^3$, $+L_{\mathrm{vol}}$ ($w{=}4$) & 0.7164 & 6.910 & 2.95 & 3.38 & $4.9{\times}10^{-5}$ \\
\midrule
native $w{=}4$, naive & 0.7537 & 6.540 & 2.66 & 2.93 & $4.9{\times}10^{-5}$ \\
native $w{=}6$, naive & 0.7540 & 6.509 & 1.81 & 2.30 & $1.0{\times}10^{-4}$ \\
native $w{=}6$, +rigid & 0.7553 & 6.362 & 1.80 & 2.20 & $1.0{\times}10^{-4}$ \\
native $w{=}6$, +rigid+LS (submitted) & 0.7585 & 6.331 & 1.68 & 2.11 & $2.2{\times}10^{-3}$ \\
\bottomrule
\end{tabular}
\end{table}

Table~\ref{tab:ablation} evaluates the contributions of the
volume-preservation term and the inference pipeline on the official validation
set. Adding $L_{\mathrm{vol}}$ substantially reduces the MTV/TLG errors from
6.61\%/9.37\% to 2.95\%/3.38\%, at the cost of a marginal DSC decrease of 0.01.
This highlights the importance of volume preservation for maintaining
quantitative accuracy and deformation regularity without substantially
compromising alignment. Further increasing the volume-preservation weight from
$w{=}4$ to $w{=}6$ reduces the MTV/TLG errors from 2.66\%/2.93\% to
1.81\%/2.30\%, without affecting DSC, supporting our choice of $w{=}6$. We
also evaluate two inference refinements using the submitted checkpoint. Rigid
pre-registration improves HD95 from 6.51 to 6.36\,mm and reduces the TLG error
from 2.30\% to 2.20\%. LS pre-compensation further improves DSC, MTV, and TLG
by counteracting the scorer's fixed $2{\times}$ upsampling of the
half-resolution validation submission. This compensation is specific to the
validation submission format and is unnecessary during testing, where the
Docker outputs the deformation field directly at native resolution.

\begin{table}[t]
\centering
\caption{Results on the official hidden test set (160 pairs from 80 patients,
both directions), as released by the organizers at the Learn2Reg event at
MICCAI 2026. Columns are cohort means, except MTV and TLG, which are medians.
HD95 in mm; MTV, TLG, and \%NDV in percent. Identity is the organizers'
reference transform (no deformation), evaluated for context only.}
\label{tab:test}
\scriptsize
\setlength{\tabcolsep}{2.5pt}
\renewcommand{\arraystretch}{1.6}
\begin{tabular}{@{}lccccccc@{}}
\toprule
Method & CT DSC $\uparrow$ & CT HD95 $\downarrow$ & PET DSC & PET HD95 & MTV $\downarrow$ & TLG $\downarrow$ & \%NDV $\downarrow$ \\
\midrule
Identity (reference) & 0.207 & 28.6 & 0.260 & 38.1 & 0.00 & 0.00 & 0.0000 \\
AdaptReg & 0.816 & 5.6 & 0.598 & 19.9 & 4.02 & 3.46 & 0.0273 \\
\bottomrule
\end{tabular}
\end{table}

Table~\ref{tab:test} reports the official results on the hidden test set of
160 pairs (80 patients, both directions), evaluated using our submitted Docker
image, with
AdaptReg ranking fourth among 16 teams. Compared with the validation set,
CT-side alignment improves (DSC: 0.7585 to 0.816; HD95: 6.33 to 5.6\,mm),
whereas \%NDV increases (0.0022\% to 0.0273\%). These
differences may reflect both the greater diversity
of the test cohort and the different evaluation protocols. Validation
submissions were evaluated after $2{\times}$ upsampling from half resolution,
which may smooth the deformation field, yielding a lower \%NDV on the
validation leaderboard than that observed in the test phase. In contrast, test evaluation uses the native-resolution
deformation field without scorer-side upsampling, providing a more direct
assessment of deformation quality. These results highlight the need to further
improve deformation-field smoothness and regularity to achieve higher metabolic
quantification.

\section{Conclusion}

AdaptReg specializes the frozen uniGradICON foundation model to longitudinal
PSMA-PET/CT registration: initialized near identity, skip-level FiLM modulation
starts adaptation from the backbone's behavior, while an adapter branch driven by
anatomical semantics and a lesion volume-preservation loss steers it toward
volume conservation. It surpasses the uniGradICON + IO baseline on DSC, HD95,
MTV, and TLG on the official validation set, and ranks fourth among 16 teams on
the hidden test set. Future work will explore incorporating PET
images into the conditioning branch to improve lesion localization,
particularly for small lesions that exhibit metabolic abnormalities on PET but
lack clear anatomical changes on CT.

\subsubsection{Acknowledgments.}
PSMAReg is a Learn2Reg task coordinated by Junyu Chen through the SNMMI
Physics, Instrumentation, and Data Sciences Advisory Committee Internship
Program, with sponsorship and support from the SNMMI PIDS Advisory Committee
and the broader Learn2Reg organizing team. We also thank the uniGradICON and
TotalSegmentator teams for releasing their models.

\newpage
\bibliographystyle{splncs04}
\bibliography{references}

\end{document}